\pdfoutput=1
\documentclass[11pt]{article}

\usepackage[]{acl}

\usepackage{times}
\usepackage{latexsym}

\usepackage[T1]{fontenc}
\usepackage[utf8]{inputenc}

\usepackage{microtype}
\usepackage{graphicx}
\usepackage{mathtools}
\usepackage{mathrsfs}
\usepackage{amsmath}
\usepackage{algorithm} 
\usepackage{algpseudocode}
\usepackage{multicol}
\usepackage{multirow}
\usepackage{caption}
\usepackage{subfigure}

\title{Adaptive Meta-learner via Gradient Similarity for \\ Few-shot Text Classification}
\author{Tianyi Lei$^1$, Honghui Hu$^1$, Qiaoyang Luo$^2$, Dezhong Peng$^1$, Xu Wang$^1$\thanks{~~Corresponding author} \\
$^1$College of Computer Science, Sichuan University \\
$^2$The University of Adelaide\\
\texttt{\{leity828,wangxu.scu\}@gmail.com}}

\begin{document}
\maketitle
\begin{abstract}
Few-shot text classification aims to classify the text under the few-shot scenario. Most of the previous methods adopt optimization-based meta learning to obtain task distribution. However, due to the neglect of matching between the few amount of samples and complicated models, as well as the distinction between useful and useless task features, these methods suffer from the overfitting issue. To address this issue, we propose a novel Adaptive Meta-learner via Gradient Similarity (AMGS) method to improve the model generalization ability to a new task. Specifically, the proposed AMGS alleviates the overfitting based on two aspects: (i) acquiring the potential semantic representation of samples and improving model generalization through the self-supervised auxiliary task in the inner loop, (ii) leveraging the adaptive meta-learner via gradient similarity to add constraints on the gradient obtained by base-learner in the outer loop. Moreover, we make a systematic analysis of the influence of regularization on the entire framework. Experimental results on several benchmarks demonstrate that the proposed AMGS consistently improves few-shot text classification performance compared with the state-of-the-art optimization-based meta-learning approaches. The code is available at: \url{https://github.com/Tianyi-Lei}.

\end{abstract}

\section{Introduction}

As a fundamental task of few-shot learning \cite{fei2006one} in natural language processing theme, few-shot text classification \cite{yu2018diverse, geng2019induction} requires a model to predict categories that are not seen in training. Meta learning \cite{schmidhuber1987evolutionary, thrun2012learning}, which plays a crucial role in general few-shot learning, aims to improve generalization ability and fast adaptation ability of the learner through modelling the distribution of tasks. To adapt few-shot tasks, typical supervised meta-learning methods \cite{vinyals2016matching, finn2017model} model task distributions from a few support tasks over meta-training episodes. Subsequently, numerous methods based on meta-learning \cite{DBLP:conf/iclr/BaoWCB20,luo2021don,han2021meta} are proposed to solve few-shot text classification problem. 

Within the meta-learning frameworks, \citet{DBLP:conf/iclr/BaoWCB20} trains an attention-based model to enhance the text representation of distributional signature, \cite{luo2021don} leverages label-semantic augmentation to help BERT compensate for the ambiguity of the class definition caused by the limited data, and \cite{han2021meta} strengthens the generalization of a model using an adversarial domain adaptation network. However, these methods are similar to the traditional meta-learning methods, neglecting the overfitting problem caused by utilizing the few number of data in the complicated models under the meta-learning frameworks.

\begin{figure*}[t]
\subfigure[MAML(First order)]
{
% 	\begin{minipage}[t]{0.4\linewidth}
	\centering
	\includegraphics[scale=0.5]{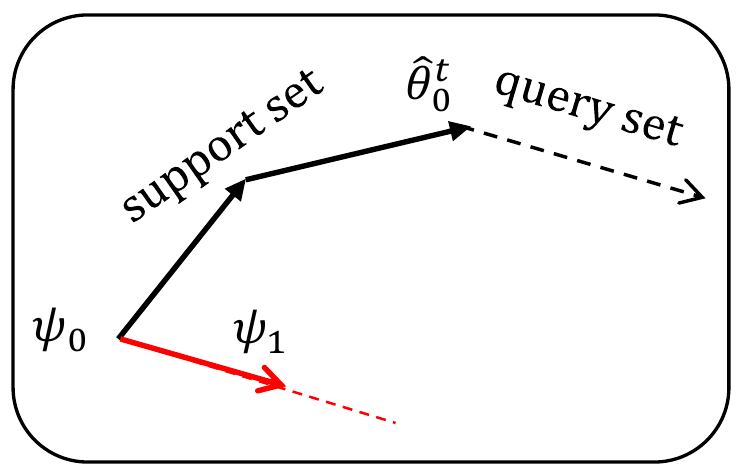} 
% 	\end{minipage}
}
% \quad
\subfigure[REPTILE]
{
% 	\begin{minipage}[t]{0.4\linewidth}
	\centering
	\includegraphics[scale=0.5]{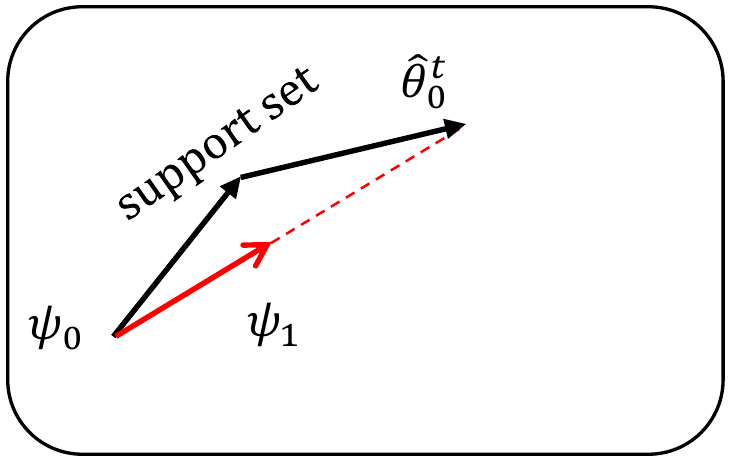} 
% 	\end{minipage}
}
% \quad
\subfigure[AMGS (ours, scheme 1)]
{
    % \begin{minipage}[t]{0.4\linewidth}
	\centering
	\includegraphics[scale=0.5]{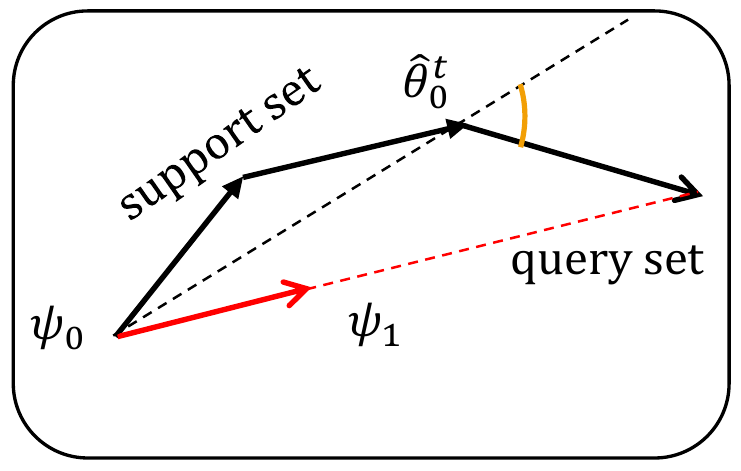}
% 	\end{minipage}
}
% \quad
\subfigure[AMGS (ours, scheme 2)]
{
    % \begin{minipage}[t]{0.4\linewidth}
	\centering
	\includegraphics[scale=0.5]{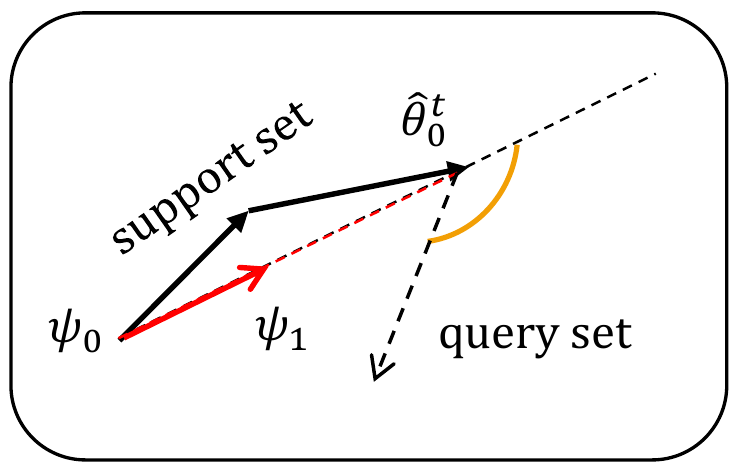}
% 	\end{minipage}
}

\caption{Diagram of the comparison of different methods for gradient direction optimization. The black arrow, black dotted arrow, red arrow and red dashed line denote the actual update of base-learner, the update direction of based-learner, the actual update of meta-learner and the update direction of meta-learner, respectively. \textbf{(a)} MAML(First order): A set of initial parameters $\psi_{0}$ is updated in the direction of the red arrow, \textit{i.e.}, the gradient of query set loss, which is calculated at $\hat\theta$ after $t$-step updates. Note that, since the gradient calculation of MAML contains the Hessian matrix, it is hard to represent in the figure, we use the First Order MAML (FOMAML) to replace MAML. \textbf{(b)} REPTILE: $\psi_{0}$ is updated along the red arrow pointing to the $t$-step optimization solution. \textbf{(c)} and \textbf{(d)} are different schemes of our proposed adaptive meta-learner, which distinguish the positive gradient cosine similarity (scheme 1) and negative gradient cosine similarity (scheme 2). If the gradient direction obtained on the query set is similar to the gradient direction of the sum of the $t$ updates (black dashed line), $\psi_{0}$ is updated in the direction of the sum of all gradients. While if their gradient directions are opposite, we remove the gradient obtained from the query set.}
\label{fig:1}
\end{figure*}

To address the above problem in few-shot text classification, several methods are proposed based on a principle \textit{i.e.}, obtaining more task-distribution can ameliorate the risk of over-fitting to the training task distribution. \citet{bansal2020self} alleviates overfitting through joint training of self-supervised tasks and classification tasks in pre-trained models. We also follow this method and use a self-supervised Mask Token Prediction (MTP) task in meta training phase. Unfortunately, the increased task distribution generated by this joint training is not always positive for meta-training.

In order to further overcome the overfitting challenge in meta-training, we propose the adaptive meta-learner via gradient similarity based on another principle \textit{i.e.}, distinguishing positive and negative features by feature selection of deep model can enhance generalization by alleviating overfitting. In optimization-based meta-learning framework, the gradient contains all the information transmitted from the inner-learner to the outer-learner, including “features” mentioned in above principle. Thus, the gradient obtained by base-learner can be regarded as the "features". To compare with other training strategies for meta-learner, we plot Figure \ref{fig:1}. Other strategies often adopt all the gradient obtained by the base-learner without distinction. They also may consume enormous computing resources for calculating the Hessian matrix, sacrifice the stability and accuracy in order to adopt the first-order algorithm, or discard the query set in the training Batch in order to simplify the calculation. By contrast, our method only needs to distinguish the gradient similarity between the gradient of the loss on the query set and the current gradient of the base-learner during the meta-training process. Subsequently, we utilize the corresponding gradient of its loss to help meta-learner quickly adapt to the optimization space. Such method selects the more useful gradients for meta-learner in current training batch. In addition, it neither increases the computational complexity nor causes waste of text information in the same training episode.

According to the above principles, we propose a novel Adaptive Meta-learner via Gradient Similarity (AMGS) algorithm based on optimization-based meta learning scheme. We firstly construct the self-supervised task called Mask Token Prediction (MTP) for the base-learner in the inner loop. Such approach can generate the extension of the task distribution from unlabeled text and constraint the gradient updating of primary classification task to increase the robustness of the model. Moreover, in the outer loop, we utilize the adaptive meta-learner to improve the utilization of the task features from the inner loop. As Figure (\ref{fig:1}) shows, our strategy can more efficiently leverage query set samples in a training episode, which optimizes the scope of gradient optimization. Therefore, the adaptive meta-learner directly accomplish additional amelioration of overfitting.

The contributions of this paper are summarized as follows: (1) We construct an optimization-based meta-learning framework named AMGS and elaborately design a meta-training algorithm to effectively tackle the overfitting issue in few-shot text classification based on two different principles. (2) We propose an adaptive meta-learner that selects the positive gradients and removes the negative gradients to improve the generalization ability of the model on the few-shot task (3) Experimental results demonstrate that the proposed AMGS outperforms the state-of-the-art optimization-based meta-learning models.

\section{Related Work}

\paragraph{Few-shot text classification via meta learning}

Few-shot learning is an application of meta-learning. In most meta-learning frameworks, the strategies can be divided into two categories: metric-based meta-learning and optimization-based meta-learning. Prototypical Network \cite{snell2017prototypical}, Induction Network \cite{geng2019induction} and Relation Network \cite{sung2018learning} are dedicated to construct a metric space between classes and samples. In the optimization-based meta-learning methods, most of them consist of an inner (or base) algorithm and an outer (or meta) algorithm. Model-Agnostic Meta-Learning (MAML) \cite{finn2017model} and Reptile \cite{DBLP:journals/corr/abs-1803-02999} are examples of such optimization-based algorithms. LEOPARD \cite{bansal2020self} achieves a good performance on diverse classification tasks by using BERT \cite{devlin2019bert}. Meanwhile, recent work \cite{DBLP:conf/iclr/BaoWCB20} proposes a meta-learning-based method by using distributional signatures for few-shot text classification. More recently, LaSAML \cite{luo2021don} uses label information for few-shot text classification. Another one \cite{han2021meta} applies a domain discriminator into a meta-learning framework. However, these algorithms suffer from overfitting caused by the imbalance between the few data and the deep model in the few-shot setting. By contrast, our proposed AMGS which expands the task distribution in the inner loop and distinguishes the positive and negative gradient in the outer loop can address this issue indirectly and directly.

\paragraph{Auxiliary learning}
In general, auxiliary learning can assist the main task to learn more accurately and quickly in deep learning \cite{wang2022dsc3l,wang2019advcae}, especially in the multi-task learning field. SSL-Reg \cite{10.1162/tacl_a_00389} builds a regularizer of the loss of self-supervised learning tasks to improve performance on text classification. Besides constructing a task, external auxiliary data can also be introduced into the model to obtain more latent information \cite{DBLP:journals/corr/abs-1807-10916}.

Similarly, auxiliary tasks are valuable to adapt the meta-learning scheme. MAXL \cite{liu2019self} adopts a self-supervised learning scheme to generate auxiliary labels, improving the generalization ability of the primary task in gradient update. Furthermore, self-supervised auxiliary tasks can promote fast adaptation during the testing phase \cite{chi2021test}. Hybrid SMLMT \cite{bansal2020self} creates a specific self-supervised auxiliary task for multi-task learning. Similar to these auxiliary tasks, our auxiliary task MTP is self-supervised to generate richer task distribution during meta-training.

\section{Methods}

In this section, we first introduce the preliminaries for few-shot classification \cite{vinyals2016matching}. Next, we describe Adaptive Meta-learner via Gradient Similarity (AMGS) method in detail.

\subsection{Overview}
\paragraph{Problem setup}
The setting of few-shot classification often includes \textit{training episode} and \textit{testing episode}. Suppose we have examples with labels from the classes $y_{train}$of training episode and need to predict the labels of examples from unseen but related classes $y_{test}$ of testing episode. The training classes and testing classes are mutually exclusive, denotes as $y_{train} \cap y_{test} = \Phi$. To create a training episode, we need to build a set of \textit{N-way K-shot} tasks. For each task, we sample $N$ classes, $k+q$ examples of each class randomly. The $N \times k$ \{$x_s, y_s$\} pairs including examples and corresponding labels constitute the support set, while the $N \times q$ labeled examples \{$x_q, y_q$\} are known as the query set. It is the same way to create a support set in testing episode, but leverage the unlabeled examples \{$y_q$\} to create the query set in testing. By repeating the above procedure, we can obtain enough training and testing episodes, so that we can use them in meta-training and meta-testing respectively. 
In short, such setting requires the model to have the ability to generalize from seen classes in training episodes to unseen classes in testing episodes.

\begin{figure}
    \centering
    \includegraphics[width=1\linewidth]{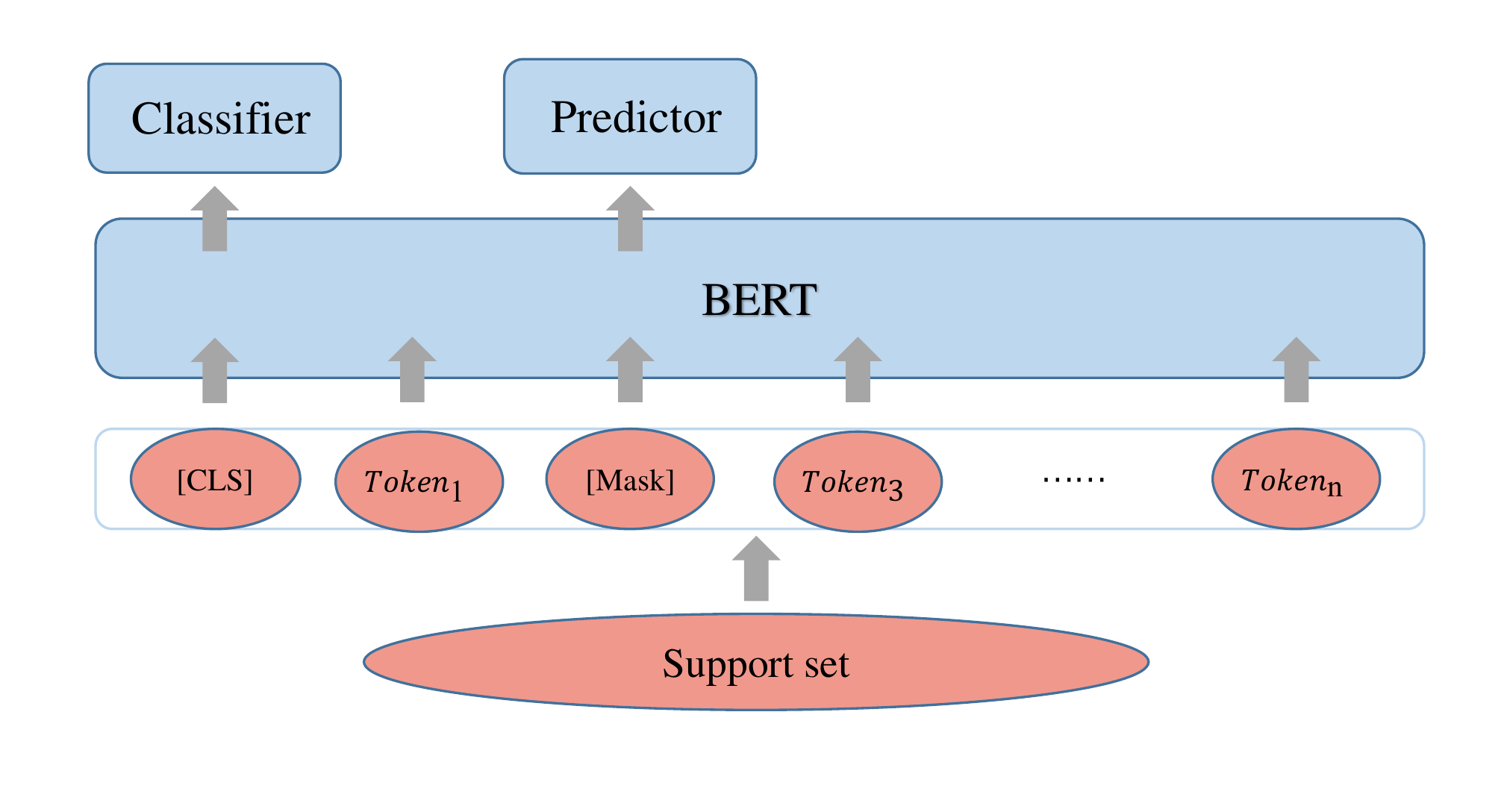}
    \caption{The main framework of the proposed AMGS.}    
    \label{fig:2}
\end{figure}

\paragraph{Model architecture} \label{model-architecture} 

BERT \cite{devlin2019bert} performs well in the conventional text classification, thus we leverage it as text encoder in our proposed AMGS framework to explore the problem of few-shot text classification. As shown in Figure (\ref{fig:2}), the model architecture consists of the BERT encoder, a classifier and a predictor. The model performs the primary task (i.e, classification) and the auxiliary task (i.e., token prediction) simultaneously, which constitutes multi-task learning. In training period, the support set are used to obtain the BERT encoding and label prediction in the primary branch. While in the auxiliary task branch, BERT encoder updates parameters through the self-supervised task without labels. 

For convenience to explain in following section, we define parameters of the total network $\theta = \{\theta_b, \theta^{pri}_c, \theta^{aux}_p\}$, where $\theta_b$ denotes the shared weights of BERT encoding, $\theta^{pri}_c$ represents classification weights for the primary task, and $\theta^{aux}_p$\ is prediction weights for the auxiliary task. Concretely, the primary-branch weights and the auxiliary-branch weights are respectively denoted as $\theta^{pri} = \{\theta_b, \theta^{pri}_c\}$, and $\theta^{aux} = \{\theta_b, \theta^{aux}_p\}$.

\paragraph{Self-supervised Mask Token Prediction task} \label{self-supervise-auxiliary-tast}

As mentioned above, we leverage BERT as text encoder. Considering that our self-supervised auxiliary task should be adapted to BERT, we adopt the Mask Token Prediction \textbf{(MTP)} task used in BERT pre-training stage (also known as MLM). MTP randomly masks the tokens in the sentences according to the specified ratio. These masked tokens are fed into BERT to be predicted by putting the final hidden vector corresponding to the masked token into the output softmax over the vocabulary. The original strategy of MTP in BERT set 15\% probability for each token to replace with the [MASK] token 80\% of the time, a random token 10\% of the time, and the unchanged token 10\% of the time. In some cases, if the text used to construct MTP task is very short, none of the tokens in this text would be masked with high probability. This will cause the effect of MTP to fail in the downstream task.

Therefore, we improve the replacement probability of each token being masked to 30\% instead of 15\%. Meanwhile, the masking rating of replacing the target token with a random token and an unchanged token are both set to 0\%, because random and unchanged replacement both occur for 3\% of all tokens, which leads to instability. This change helps the model acquire the new task distribution more stably. The masked strategy is explored in experiment demonstrated in Appendix \ref{Additional Ablation Studies}.

\subsection{Adaptive Meta-learner via Gradient Similarity (AMGS)}
The optimization-based meta-learning methods \cite{finn2017model,DBLP:journals/corr/abs-1803-02999} learn an appropriate initial parameters by meta learner, achieving encouraging performance. However, these methods ignore the overfitting issue in the few-shot learning. Considering that the direction of gradients could be used to distinguish the positive and negative gradients, we propose AMGS framework with explicit regularization. The training procedure of AMGS is decomposed into two steps: (i) The base-learner collects gradient for adaptive meta-learner, which utilizes the multi-task network to learn primary and auxiliary tasks together on support set. Then it collects the gradient of the loss on the query set by leveraging the supervised primary task. (ii) The adaptive meta-learner via gradient similarity distinguishes the positive and negative gradient obtained by the first stage, then updates the parameters of the total meta-network by meta-learner. By completing two training steps, our method ensures that the meta-learner learns the more balanced initial parameters and makes the loss of new tasks decrease faster.

\begin{algorithm*}	
    \caption{Training procedure of AMGS}
	\label{al1}
	\begin{algorithmic}[0] 
	\Require learning rate $\alpha$, $\beta$, texts and corresponding labels $x, y$
	\end{algorithmic}
    Initialize $\Psi = \theta = \{\theta_b, \theta^{pri}_c, \theta^{aux}_p\}$ with BERT
    \begin{algorithmic}[1]
    \While {not converged}
        \State Sample batch of tasks ${T_i}\sim{p(T)}$
        \State Sample support set ($x_{s}$, $y_{s}$), query set ($x_{q}$, $y_{q}$)
        \For {all $T_i$}
            \State Compute adapted parameters with gradient descents: $\hat\theta = \theta - \alpha \nabla_\theta \mathcal{L}^{total}_{T_i}(x_{s}, y_{s}; \theta)$
            \State Compute the gradients of primary task on $\hat\theta$: $(\theta_b, \theta^{pri}_c) = (\theta_b, \theta^{pri}_c) - \alpha \nabla_\theta \mathcal{L}^{pri}_{T_i}(x_{q}, y_{q}; \hat\theta_b, \hat\theta^{pri}_c)$
        \EndFor
        \If {$cos( \nabla_\theta\mathcal{L}^{total}_{T_i}(x_{s}, y_{s}; \theta), \nabla_\theta \mathcal{L}^{pri}_{T_i}(x_{q}, y_{q}; \hat\theta_b, \hat\theta^{pri}_c))\geq 0$}
           \State Update: $\hat\Psi \leftarrow \Psi - \beta \nabla_\Psi \sum_{{T_i}\sim{p(T)}} (\mathcal{L}^{total}_{T_i}(x_{s}, y_{s}; \theta) + \mathcal{L}^{pri}_{T_i}(x_{q}, y_{q}; \hat\theta^{pri}))$
       \Else
           \State Update: $\hat\Psi \leftarrow \Psi - \beta \nabla_\Psi \sum_{{T_i}\sim{p(T)}} \mathcal{L}^{total}_{T_i}(x_{s}, y_{s}; \theta)$
        \EndIf
    \EndWhile
    \end{algorithmic} 
\end{algorithm*}

\subsubsection{Collecting gradient for adaptive meta-learner} \label{MTP}

This subsection describes that how the base-learner collects gradient for adaptive meta-learner. We leverage the self-supervised MTP task to acquire a more abundant task distribution and improve the base-learner robustness. In addition, as mentioned above, we build multi-task learning by using the MTP to limit the training of classification tasks. In other words, the constraint on the loss of the primary task has been enforced via the auxiliary task. This limitation prevents the base-learner to obtain extra characteristics of each training task to alleviate overfitting. Formally, we compute the total loss of the multi-task network as follows:
\begin{equation} \label{eq1}
    \mathcal{L}_{total}=(1-\rho)\mathcal{L}_{pri} + \rho\mathcal{L}_{aux},
\end{equation}

where $\mathcal{L}_{total}$, $\mathcal{L}_{pri}$, $\mathcal{L}_{aux}$ and $\rho$ represent the total loss, primary classification loss $\mathcal{L}_{pri}(x, y; \theta^{pri})$, auxiliary prediction task loss $\mathcal{L}_{aux}(x; \theta^{aux})$, and the contribution of the auxiliary task, respectively. $x$ and $y$ denote training texts and their labels. We use cross entropy loss to implement both text classification and the masked token prediction. In our experiments, we set $\rho=10^{-3}$. The sensitivity study is shown in Appendix \ref{trade-off}.

When training on the tasks $T_i$ in the \textit{support set}, the total loss Eq.(\ref{eq1}) after one or a few gradient updates can be defined as follows:
\begin{equation} \label{eq2}
    \hat\theta = \theta - \alpha \nabla_\theta \mathcal{L}^{total}_{T_i}(x_{s}, y_{s}; \theta),
\end{equation}
where $x_{s}$ and $y_{s}$ are texts and corresponding labels in the support set. $\alpha$ is the adaptation learning rate. By Eq.(\ref{eq1}) and Eq.(\ref{eq2}), we can obtain more semantic representation to apply explicit regularization to the primary loss.

In general, the query set is used for testing and inference, while it contains rich task distribution which can be applied to meta-learn. We argue that the query set can be used to fine-tune and enhance the gradient learned by the base-learner through the multi-task network. In the step, we accomplish the collection of gradient of the parameters \{$\theta_b$, $\theta^{pri}_c$\} on the query set. Finally, the objective can be defined as follows:
\begin{equation} \label{eq3}
    \underset{\theta_b, \theta^{pri}_c}{\arg\min}\ \mathcal{L}^{pri}_{T_i}(x_{q}, y_{q}; \hat\theta^{pri}),
\end{equation}
where $x_{q}$ and $y_{q}$ are texts and corresponding labels in the query set. 

\subsubsection{Upgrade meta-learner with AMGS}
This stage is mainly about updating meta-learner. Following previous work \cite{du2018adapting}, we leverage the gradient cosine similarity to measure whether the gradients obtained on query set are positive or negative. Based on Eq.(\ref{eq2}) and Eq.(\ref{eq3}), we get the gradient cosine similarity by calculating $cos( \nabla_\theta\mathcal{L}^{total}_{T_i}(x_{s}, y_{s}; \theta), \nabla_\theta \mathcal{L}^{pri}_{T_i}(x_{q}, y_{q}; \hat\theta_b, \hat\theta^{pri}_c))$. If the value of $cos(\cdot)$ is non-negative, such gradient is regarded as the \textbf{positive gradient}, which means the query set at this batch is beneficial to enhance generalization of the model. Therefore we obtain the gradient of its loss to perform gradient enhancement on the meta-learner. For this situation, the meta-objective can be written as: 
\begin{equation}
\begin{aligned}\label{eq4}
    & \underset{\theta_b, \theta^{pri}_c, \theta^{aux}_p}{\arg\min} \sum_{{T_i}\sim{p(T)}} (\mathcal{L}^{total}_{T_i}(x_{s}, y_{s}; \theta) \\
    & + \mathcal{L}^{pri}_{T_i}(x_{q}, y_{q}; \hat\theta^{pri})).
\end{aligned}
\end{equation}

On the contrary, if $cos(\cdot)$ is negative, such gradient is considered as the \textbf{negative gradient}. We remove this query set loss to ensure that the model is not negatively affected, so the meta-objective is:
\begin{equation}
\begin{aligned}\label{eq5}
    & \underset{\theta_b, \theta^{pri}_c, \theta^{aux}_p}{\arg\min} \sum_{{T_i}\sim{p(T)}} (\mathcal{L}^{total}_{T_i}(x_{s}, y_{s}; \theta)).
\end{aligned}
\end{equation}
According to above training procedure, our proposed meta-objective can distinguish the positive to use and the negative to by adaptive meta-learner, which can automatically filter appropriate regularization to limit the gradient optimization. This step reduces effective model capacity, hence it effectively alleviates overfitting and improves the generalization ability of the model. The full training procedure is demonstrated in the Algorithm \ref{al1}.

\subsubsection{Meta testing}
The model parameters have been learned in meta training phase, and fine-tuned in the meta-learning testing phase for downstream tasks. MTP can continue to participate in the fine-tuning phase in order to help the primary classification adapt to the unseen classes for the new tasks quickly. From the perspective of test-time fast adaptation \cite{chi2021test}, our auxiliary task boosts the fast gradient descent of the loss function of the primary task in the testing procedure.

\begin{table*}
\renewcommand\arraystretch{1.5}
\centering
\resizebox{\textwidth}{!}{
\begin{tabular}{lc|cc|cccc|cccc}
\hline 
\multicolumn{2}{c|}{\multirow{3}*{\textbf{Methods}}} & \multicolumn{2}{c|}{\textbf{HuffPost}} & \multicolumn{4}{c|}{\textbf{Banking77}} & \multicolumn{4}{c}{\textbf{Clinc150} (cross domain)} \\ 
\cline{3-12}
& & \multicolumn{2}{c|}{5-way} & \multicolumn{2}{c}{10-way} & \multicolumn{2}{c|}{15-way} & \multicolumn{2}{c}{10-way} & \multicolumn{2}{c}{15-way} \\
& & 1-shot & 5-shot & 1-shot & 5-shot & 1-shot & 5-shot & 1-shot & 5-shot & 1-shot & 5-shot \\
\hline
\multirow{3}*{\textbf{Metric}} & BERT+PROTO & $40.59$ & $53.48$ & $63.0$5 & $78.60$ & $59.18$ & $74.12$ & $57.43$ & $72.90$ & $52.31$ & $66.06$ \\
& BERT+RELATION & $40.80$ & $51.87$ & $63.88$ & $73.48$ & $56.29$ & $64.57$ & $54.65$ & $60.09$ & $46.54$ & $58.83$ \\
& BERT+INDUCT & $39.96$ & $50.79$ & $48.72$ & $64.32$ & $49.45$ & $55.27$ & $46.52$ & $57.65$ & $41.72$ & $49.98$ \\
\hline
\multirow{6}*{\textbf{Optimization}} & BERT+MAML & $41.03$ & $57.13$ & $59.21$ & $85.55$ & $55.69$ & $81.48$ & $60.14$ & $80.24$ & $55.00$ & $65.20$\\
& BERT+REPTILE & $40.80$ & $58.96$ & $58.36$ & $82.81$ & $56.69$ & $81.14$ & $59.89$ & $81.23$ & $53.32$ & $63.04$ \\
& BERT+R2D2 & $40.78$ & $61.98$ & $70.45$ & $87.80$ & $63.46$ & $\textbf{85.65}$ & $62.72$ & $87.13$ & $57.61$ & $80.76$ \\
& DS+R2D2 & $41.34$ & $62.48$ & $59.33$ & $83.71$ & $53.37$ & $78.96$ & $55.56$ & $78.76$ & $53.41$ & $79.69$ \\
& MLADA+R2D2 & $41.55$ & $59.82$ & $61.69$ & $80.81$ & $55.63$ & $74.77$ & $65.28$ & $85.45$ & $51.76$ & $77.77$ \\
& \textbf{BERT+AMGS (OURS)} & $\textbf{43.47}$ & $\textbf{63.40}$ & $\textbf{71.41}$ & $\textbf{88.81}$ & $\textbf{63.62}$ & ${84.93}$ & $\textbf{69.19}$ & $\textbf{88.26}$ & $\textbf{62.12}$ & $\textbf{84.13}$ \\
\hline
\end{tabular}}
\caption{\label{tb1} Results of 5-way 1-shot and 5-way 5-shot on HuffPost headlines dataset, 10-way 1-shot, 10-way 5-shot, 15-way 1-shot and 15-way 5-shot on Banking77 and Clinc150 datasets (cross domain) by using our proposed method and all baselines.}
\end{table*}

\section{Experiments}
\label{sec:length}

\subsection{Datasets}

We use three datasets to evaluate the performance in experiment.

\textbf{HuffPost headlines} includes 36900 news headlines among 41 classes, which contains less information than other datasets. In order to complete a fair comparison test, we divide each training, validation, and testing set into 20, 5, and 16 classes by following the setting of \cite{DBLP:conf/iclr/BaoWCB20}. \textbf{Banking77} \cite{casanueva2020efficient} consists of 13083 fine-grained intents and 77 classes. As for the setting of data distribution and \textit{N-way K-shot} classification tasks, we assign 30, 15, and 32 classes fixedly for training, validation, and testing set. \textbf{Clinc150} \cite{wang2019glue} is a cross-domain intent classification dataset with 150 classes in 10 domains. It provides 22500 examples that cover 150 intents from 10 domains without overlap among classes. We allocate for each training, validation, and testing with 4, 1, 5 domains, respectively.

\subsection{Baselines}
In order to evaluate our AMGS, we compare with three metric-based methods and five optimization-based algorithms for few-shot text classification.

\textbf{Proto} \citep{snell2017prototypical} provides a metric-based method to learn the class vector by computing distances to prototype representations of each class. \textbf{Induct} \cite{geng2019induction} learns a generalized class-wise representation by leveraging the dynamic routing algorithm. \textbf{Relation} \cite{sung2018learning} compares the class vector and the query feature through a relation-based meta-learner.
\textbf{MAML} \cite{finn2017model} is one of the most typical optimization-based meta-learning algorithms, which trains a favorable initial point for the base learner by utilizing the meta learning that learns among tasks. \textbf{Reptile} \cite{DBLP:journals/corr/abs-1803-02999} is a first-order variant method of MAML. It achieves that the speed of calculation is greatly improved and the complexity is reduced, while the accuracy is almost the same as MAML. The base learner used by Ridge Regression Differentiable Discriminator \textbf{(R2D2)} \cite{DBLP:conf/iclr/BertinettoHTV19} is ridge regression based on linear regression model. The amount of calculation is related to the sample size of the task, which is conducive to the learning of the meta learner. \textbf{DS} \cite{DBLP:conf/iclr/BaoWCB20} shows the best performance by leveraging the model that builds an attention generator and a ridge regressor to enhance the representational power of distributional signature. \textbf{MLADA} \cite{han2021meta} uses the meta-learning adversarial domain adaptation network to improve the adaptation and new classes embedding generation by creating a domain discriminator.

\subsection{Implementation details}
$BERT_{base}$ is used as the text encoder of all baselines. Because DS and MLADA have special requirements for textual representation and feature extraction, forcibly using BERT as encoder will be counterproductive. Thus, we re-implement the pre-train fastText embeddings \cite{joulin2016fasttext} for those model, and follow other settings in the original papers \cite{DBLP:conf/iclr/BaoWCB20, han2021meta}. For the sake of fairness, the classifiers of these two algorithms use R2D2, so we constructed a comparison item with BERT as encoder.

All parameters are optimized with Adam optimizer \cite{DBLP:journals/corr/KingmaB14}. The initial learning rates $\alpha, \beta$ are separately set to $5e-5$ and $2e-5$, and we utilize 5 gradient updates for the base adaptation step. As for the \textit{N-way K-shot} classification setting, all experiments use 25 examples for the query set. We randomly sample 100 training episodes, 100 validation episodes, and 1000 testing episodes per epoch and apply early stopping on validation for 20 epochs. We evaluate the performance of the model based on 5 different random seeds. All experiments are conducted on a GEFORCE RTX 3090 GPU.

\subsection{Experimental results}
\begin{figure}[ht]
\subfigure[AMGS (OURS)] 
{
	\begin{minipage}[t]{0.45\linewidth}
	\centering
	\includegraphics[scale=0.45]{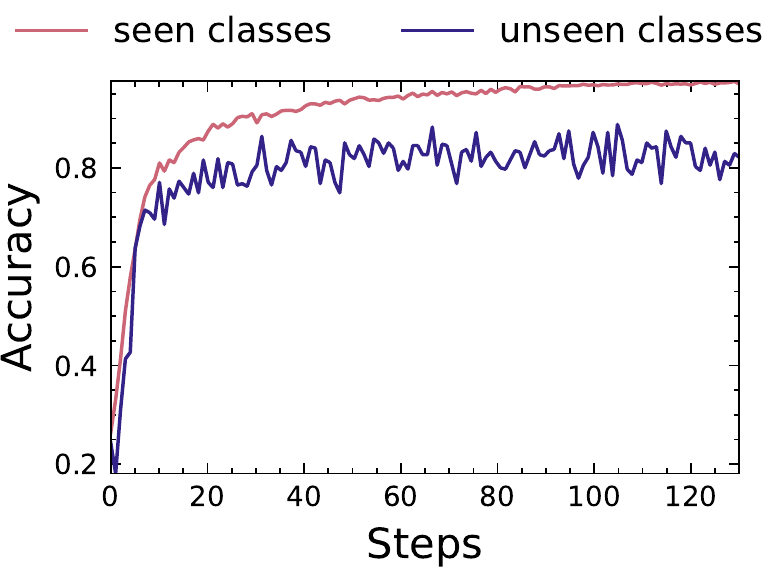} 
	\end{minipage}
}
\subfigure[MAML]
{
	\begin{minipage}[t]{0.45\linewidth}
	\centering 
	\includegraphics[scale=0.45]{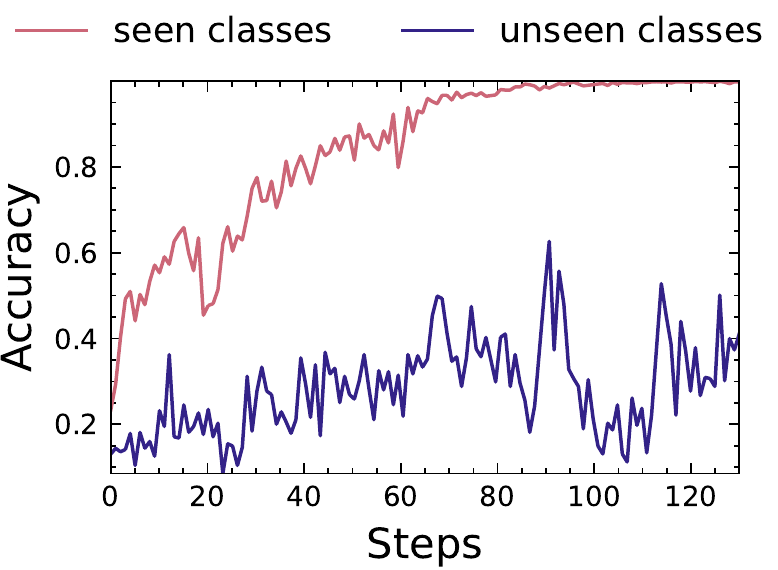}  
	\end{minipage}
}
\caption{Learning curves of AMGS (a) and MAML (b) on 15-way 5-shot task of the Banking77 dataset. We plot average accuracy from seen classes (red) and unseen classes (blue).}
\label{fig:3}
\end{figure}

The total results of experiments are reported in table \ref{tb1}. By observing these experimental results, we obtain the following conclusions: 

(1) Whether it is for texts with minimal semantics (Huffpost), fine-grained categorized (Banking77) or cross domains (Clinc150), our proposed method AMGS has an average improvement of 0.2-6.5\% over the state-of-the-art model on both 1-shot and 5-shot classification. In particular, compared with our AMGS and MAML \cite{finn2017model}, Reptile \cite{DBLP:journals/corr/abs-1803-02999}, we can draw the following observations from the Table \ref{tb1}: (i) Our proposed method achieves better performance on all tasks. In especial, in the 15-Way 5-shot task on Clinc150 dataset, our proposed method outperforms the best counterpart by 18.9\%. (ii) MAML and Reptile perform better on fine-grained classification Banking77 dataset with more similar categories than on cross-domain Clinc150 dataset with less similar categories, and have a smaller gap with our AMGS. To verify that our AMGS perform better than MAML on alleviating overfitting, we plot their accuracy learning curves in Figure \ref{fig:3}. In the figure, the training procudure of our AMGS is more stable than that of MAML from the beginning to the end. Besides, the gap between the accuracy of seen classes and unseen classes of our AMGS is less than that of MAML. These results are demonstrated that our AMGS can make model more stable in meta-training and more readily generalizeds to unseen classes by addressing the overfitting issue.

(2) With leveraging BERT as our text encoder, our method is better than all compared methods on Huffpost dataset. In \cite{DBLP:conf/iclr/BaoWCB20}, it points out that BERT can better deal with highly contextual classification but not the keyword-based news classification, e.g., Huffpost dataset. Thus, "DS+R2D2" performs better on Huffpost than "BERT+R2D2", but worse on Banking77 and Clinc150. Nonetheless, our "BERT+AMGS" surpasses all BERT-based and non-BERT-based approaches on Huffpost dataset, which shows the superiority of our AMGS method. Furthermore, the performance of our model is increased by 2.1\% on 1-shot classification and 0.9\% on 5-shot classification when compared with BERT-based models.

Overall, the above observations point that AMGS can learn the commonalities and characteristics between few-shot task distribution well by mitigating overfitting, thereby obtaining a better initialized parameter for fast adaptation.

\begin{figure*}[ht]
\subfigure[AMGS (OURS)] 
{
	\begin{minipage}[t]{0.5\linewidth}
	\centering
	\includegraphics[scale=0.5]{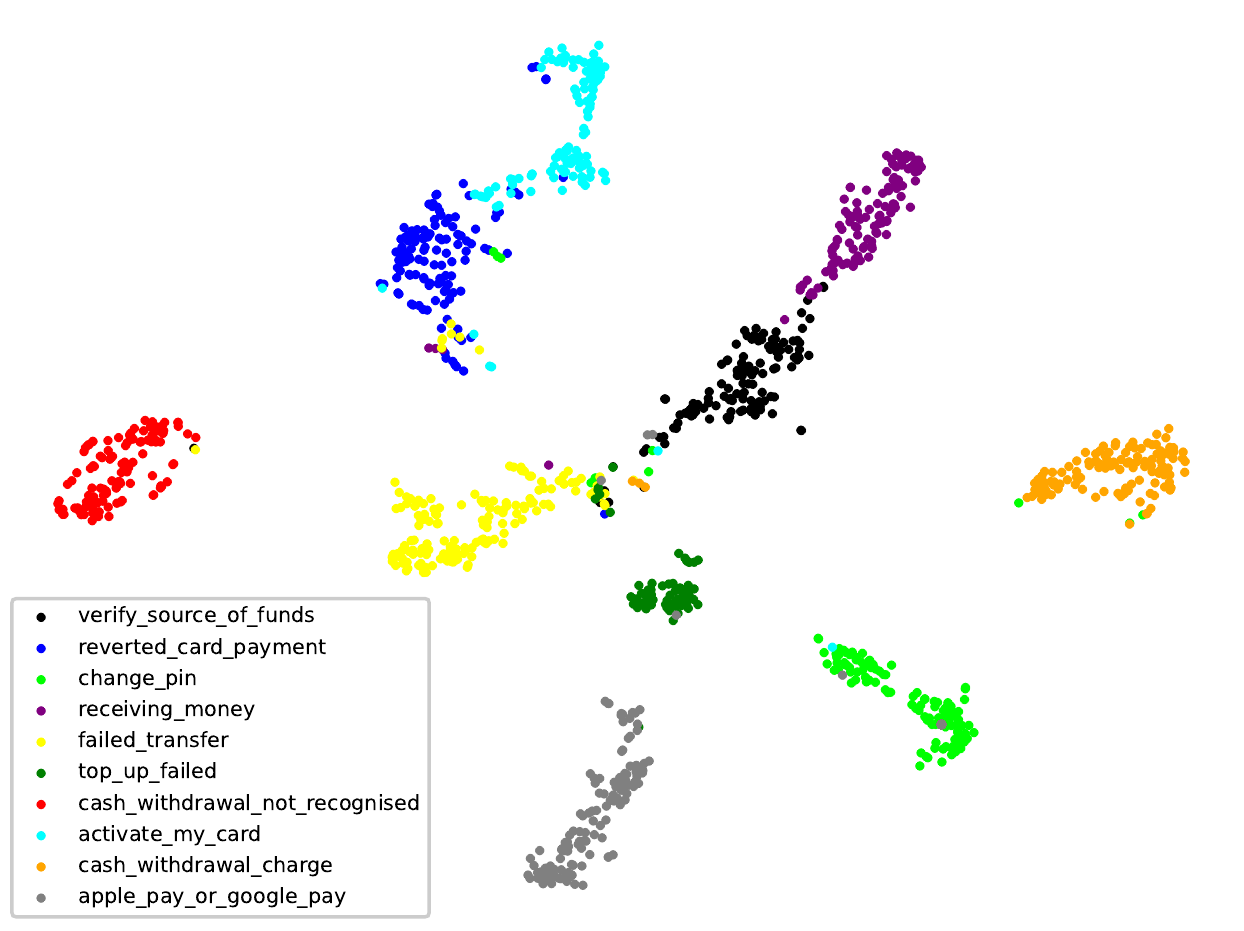} 
	\end{minipage}
}
\subfigure[REPTILE]
{
	\begin{minipage}[t]{0.5\linewidth}
	\centering 
	\includegraphics[scale=0.5]{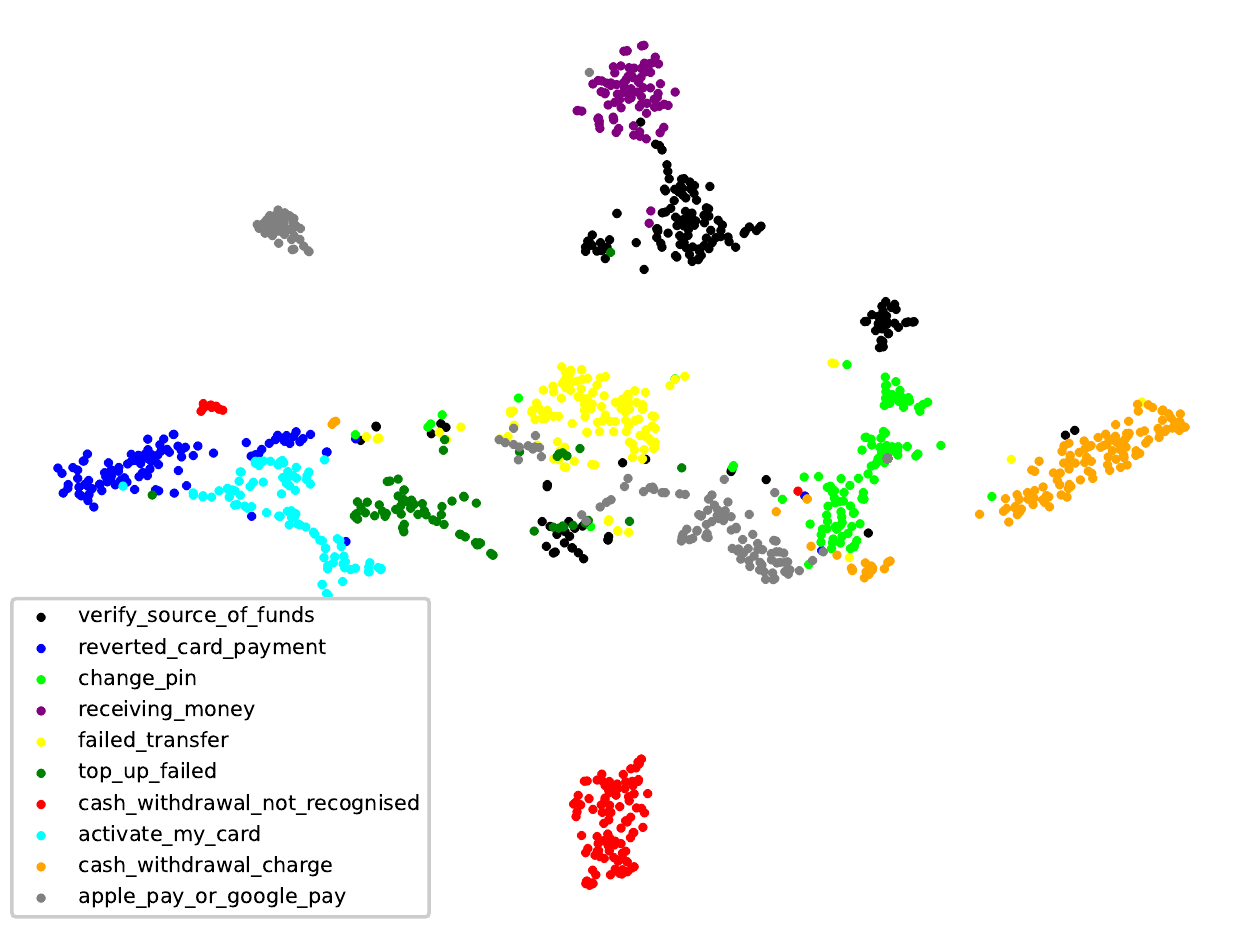}  
	\end{minipage}
}
\caption{t-SNE visualization of the input representation for the query set of a testing episode (\textit{N}=10, \textit{K}=5, \textit{L}=120) sampled from Banking77 dataset.}
\label{fig:4}
\end{figure*}

\subsection{Ablation studies}

In this section, we conduct several ablation experiments to verify the effectiveness of the adaptive meta-learner, MTP in the meta training phase, and MTP in the meta-testing fast adaptation phase. 

\begin{table}
\renewcommand\arraystretch{1.5}
\small
\centering
\scalebox{0.9}{
\begin{tabular}{lcccc}
\hline 
\multirow{3}*{\textbf{Methods}} & \multicolumn{2}{c}{\textbf{Banking77}} & \multicolumn{2}{c}{\textbf{Clinc150}} \\ 
\cline{2-5}
 & \multicolumn{2}{c}{15-way} & \multicolumn{2}{c}{15-way}\\ \cline{2-5}
 & 1-shot & 5-shot & 1-shot & 5-shot\\
\hline
AMGS w que & $56.39$ & $82.32$ & $55.72$ & $65.61$\\
AMGS w sup & $57.03$ & $81.84$ & $54.17$ & $64.12$\\
AMGS w que+sup & $58.51$ & $82.10$ & $55.97$ & $83.73$\\
AMGS w our strategy & $\textbf{63.62}$ & $\textbf{84.93}$ & $\textbf{62.12}$ & $\textbf{84.13}$\\
\hline
\end{tabular}}
\caption{\label{tb2} Ablation study results for different strategies of meta-learner on Banking77 and Clinc150 (cross domain) datasets.}
\end{table}
\paragraph{The effectiveness of the adaptive meta-learner} In this section, we further investigate the impact of the different strategies for meta-learner. To compare with our strategy, we design three other comparison strategies. As shown in Table \ref{tb2}, "AMGS w que", "AMGS w sup", "AMGS w que+sup" respectively represent the meta-learner in AMGS only use the gradients of the query set, support set and both query and support set. None of these three strategies pay attention to distinguishing the positive or the negative of the gradients. Comparing our strategy with "AMGS w que+sup" strategy, we have improved significantly more on 1-shot task than on 5-shot task. From all the results, our adaptive meta-learner which filters the impact of the negative gradient achieves the better performance among these compared strategies. 
\begin{table}
\renewcommand\arraystretch{1.5}
\small
\centering
\scalebox{0.9}{
\begin{tabular}{lcccc}
\hline 
\multirow{3}*{\textbf{Methods}} & \multicolumn{2}{c}{\textbf{Banking77}} & \multicolumn{2}{c}{\textbf{Clinc150}} \\ 
\cline{2-5}
 & \multicolumn{2}{c}{15-way} & \multicolumn{2}{c}{15-way}\\ \cline{2-5}
 & 1-shot & 5-shot & 1-shot & 5-shot\\
\hline
AMGS w/o MTP & $62.82$ & $84.73$ & $61.82$ & $83.17$\\
AMGS w/o MTP (testing) & $63.26$ & $84.32$ & $61.97$ & $84.01$\\
AMGS w MTP & $\textbf{63.62}$ & $\textbf{84.93}$ & $\textbf{62.12}$ & $\textbf{84.13}$\\
\hline
\end{tabular}}
\caption{\label{tb3} Ablation study results of MTP in meta-training and meta-testing fast adaptation phase (testing) on Banking77 and Clinc150 (cross domain) datasets.}
\end{table}
\paragraph{The effectiveness of MTP in meta-training phase and in meta-testing fast adaptation phase} As shown in Table \ref{tb3}, we first eliminate MTP in training stage. After losing a richer distribution of tasks, the performances of AMGS decrease by about 0.8\%, which verifies the effectiveness of MTP in meta-training phase. Further, we explore MTP in meta-testing fast adaptation phase. The empirical results demonstrate that after joining the auxiliary task in meta-testing, the model performances have increased by about 0.5\%. The testing auxiliary task makes the primary task more robust on the support set, and has some suppression effects on the occurrence of overfitting. All these results demonstrate that MTP task have a certain effect on Banking77 and Clinc150 datasets, but it can not significantly improve the experimental results.

% All these results prove that MTP used to joint training with classification task plays an indelible role in both meta-training and meta-testing fast adaptation phase. 

\subsection{Visualization}

We visualize the results of the experiments to demonstrate that our model can generate a high-quality text representation for unseen classes.

T-SNE \cite{van2008visualizing} visualization illustrates the experimental results in Figure (\ref{fig:4}), we take out the generated sentence embedding layer before sending it to the classifier for visualization. Comparing Figure \ref{fig:4}(a) and Figure \ref{fig:4}(b), it is obvious that our method AMGS produces better separation than REPTILE, Especially for the categories represented by gray and lime, the sentence representations obtained by REPTILE are very similar, so that it is difficult to distinguish their categories. The above observations demonstrate the effectiveness of AMGS to generate a high-quality text representation for few-shot text classification.

% Especially for the categories represented by gray and lime, the sentence representations obtained by REPTILE are very similar, so that it is difficult to distinguish their categories. However, the between-class of these two classes in the representation of RMAL is very large. This is precisely the difficulty of Banking77 dataset because its categories are very similar and even have some fine-grained classification situations.

\section{Conclusion}

In this paper, we present an Adaptive Meta-learner via Gradient Similarity (AMGS) framework for few-shot text classification. To be specific, we first leverage the self-supervised Mask Token Prediction (MTP) task to enrich the task distribution with the unlabeled text. Such approach can reduce the impact of overfitting caused by the mismatching between the few samples and the deep model. Secondly, we construct an adaptive meta-learner via gradient similarity for the outer loop to distinguish the positive and negative gradient. Thus, the meta-learner alleviates overfitting by preventing the influence of negative features. Experimental results validate that our model achieves significant improvement on the few-shot text classification tasks by effectively alleviating the overfitting issue.

\section{Acknowledge}
This work is supported by the National Natural Science Foundation of China (Grant No. U19A2078), China Postdoctoral Science Foundation (Nos. 2021TQ0223, 2022M712236), Sichuan Science and Technology Planning Project (Grant Nos. 2022YFQ0014, 2022YFH0021, 2021YFS0390), and Fundamental Research Funds for the Central Universities (No. 2022SCU12081).

\bibliography{custom}
\bibliographystyle{acl_natbib}

\appendix

\section{Ablation for different MTP masking strategies}\label{Additional Ablation Studies}

In section \ref{self-supervise-auxiliary-tast}, we mention that MTP adopts a different masking strategy from the one used in BERT pre-train stage. We explore the effect of different masking strategies in following ablation.

\begin{table}[H]
\renewcommand\arraystretch{1.5}
\small
\centering
\scalebox{0.9}{
\begin{tabular}{c|ccc|c}
\hline 
\multirow{2}*{\textbf{Masking prob.}} & \multicolumn{3}{c|}{\textbf{Masking strategy}} & \textbf{Banking77} \\
\cline{2-4}
 & Mask & Same & Random & 10-way 1-shot \\
\hline
\multirow{2}*{15\%} & 100\% & 0\% & 0\% & 73.38 \\
 & 80\% & 10\% & 10\% & 72.06 \\
\multirow{2}*{\textbf{30\%}} & \textbf{100\%} & \textbf{0\%} & \textbf{0\%} & \textbf{74.06} \\
 & 80\% & 10\% & 10\% & 72.20 \\
\multirow{2}*{45\%} & 100\% & 0\% & 0\% & 72.56 \\
 & 80\% & 10\% & 10\% & 72.42 \\
\hline
\end{tabular}}
\caption{\label{tb4} Ablation study results of different masking strategies on the validation episodes of Banking77.}
\end{table}

As Table \ref{tb3} shows, we explore the effect of different masking probabilities and strategies on Banking77. In the table, "Mask" means that we replace the token with [MASK] in MTP, "Same" means that we keep the target token unchanged and "Random" means that the token is replaced with the random token except itself. From the table, we can see that the masking strategy in BERT pre-training is not the best choice in the few-shot text classification. Therefore, in this paper, we attempt to alter the masking strategy which 100\% changes the target token to [MASK].

\section{Sensitivity study on the trade-off parameter $\rho$} \label{trade-off}
In order to set an appropriate value for the trade-off parameter of MTP mentioned in section \ref{MTP}, we study a sensitivity study for this hyper-parameter in 10-way 5-shot on Banking77 dataset.

\begin{table}[H]
\renewcommand\arraystretch{1.5}
\small
\centering
\scalebox{0.9}{
\begin{tabular}{c|cccccc}
\hline 
trade-off $\rho$ & $0.9$ & $0.5$ & $10^{-1}$ & $10^{-3}$ & $10^{-5}$ & $0$ \\
\hline
Accuracy & 89.98 & 90.10 & 94.80 & \textbf{95.60} & 94.00 & 93.40\\
\hline
\end{tabular}}
\caption{\label{tb5} Sensitivity study results of 10-way 5-shot on the validation episodes of Banking77.}
\end{table}

The results of validation episodes have shown in Table \ref{tb4}. We explore a large scale trade-off $\rho$. demonstrating MTP has the greatest contribution when the trade-off $\rho$ equals 0.001. Especially, $\rho$ equals 0 means we remove the impact of MTP, which verifies the effectiveness of our MTP.

% \section{Datasets} \label{datasets}
% Due to the particularity of the few-shot classification, the sampling of the task is very strict.  In addition to ensuring that the random seeds are the same in the comparative experiment, the division of the training, validation and testing set also needs to be consistent. 
% \paragraph{HuffPost headlines} In order to complete a fair comparison test, we divide each training, validation, and testing set into 20, 5, and 16 classes by following the setting of \cite{DBLP:conf/iclr/BaoWCB20}.
% \paragraph{Banking77} As for the setting of data distribution and \textit{N-way K-shot} classification tasks, we assign 30, 15, and 32 classes fixedly for training, validation, and testing set.
% \paragraph{Clinc150} It provides 22500 examples that cover 150 intents from 10 domains without overlap among classes. We allocate for each training, validation, and testing with 4, 1, 5 domains, respectively. 

\end{document}